\documentclass[10pt,twocolumn,letterpaper]{article}

\usepackage[pagenumbers]{cvpr} 

\usepackage{graphicx}
\usepackage{amsmath}
\usepackage{amssymb}
\usepackage{booktabs}
\usepackage[ruled]{algorithm2e}
\usepackage{multirow}
\usepackage{epsfig}
\usepackage[misc]{ifsym}

\usepackage[pagebackref,breaklinks,colorlinks]{hyperref}

\usepackage[capitalize]{cleveref}
\crefname{section}{Sec.}{Secs.}
\Crefname{section}{Section}{Sections}
\Crefname{table}{Table}{Tables}
\crefname{table}{Tab.}{Tabs.}

\def\cvprPaperID{***} 
\def\confName{CVPR}
\def\confYear{2022}

\begin{document}

\title{Rethinking Depth Estimation for Multi-View Stereo: A Unified Representation}

\author{Rui Peng \textsuperscript{1} \quad Rongjie Wang \textsuperscript{2} \quad Zhenyu Wang \textsuperscript{1} \quad Yawen Lai \textsuperscript{1} \quad Ronggang Wang \textsuperscript{\Letter,1,2}\\
\textsuperscript{1}School of Electronic and Computer Engineering, Peking University \quad \textsuperscript{2}Peng Cheng Laboratory\\
{\tt\small ruipeng@stu.pku.edu.cn \quad rgwang@pkusz.edu.cn} \\
{ \url{https://github.com/prstrive/UniMVSNet}}
}

\maketitle

\begin{abstract}
  Depth estimation is solved as a regression or classification problem in existing 
  learning-based multi-view stereo methods. Although these two representations have 
  recently demonstrated their excellent performance, they still have apparent 
  shortcomings, \eg, regression methods tend to overfit due to the indirect learning 
  cost volume, and classification methods cannot directly infer the exact depth due 
  to its discrete prediction. In this paper, we propose a novel representation, termed 
  \textbf{Unification}, to unify the advantages of regression and classification. It can 
  directly constrain the cost volume like classification methods, but also realize 
  the sub-pixel depth prediction like regression methods. To excavate the potential of 
  unification, we design a new loss function named \textbf{Unified Focal Loss},
  which is more uniform and reasonable to combat the challenge of sample imbalance.
  Combining these two 
  unburdened modules, we present a coarse-to-fine framework, that we call 
  \textbf{UniMVSNet}. The results of ranking \textbf{first} on both DTU and Tanks and Temples 
  benchmarks verify that our model not only performs the best but also has the best 
  generalization ability.
\end{abstract}

\section{Introduction}
\label{sec:intro}

Multi-view stereo (MVS) is a vital branch to extract geometry from photographs, 
which takes stereo correspondence from multiple images as the main cue to 
reconstruct dense 3D representations. Although traditional methods 
\cite{seitz2006comparison, barnes2009patchmatch,
furukawa2009accurate,schonberger2016pixelwise} have achieved excellent 
performance after occupying researchers for decades, more and more learning-based 
approaches \cite{yao2018mvsnet,yao2019recurrent,chen2019point,cheng2020deep,
gu2020cascade,yang2020cost} are proposed to promote the effectiveness of 
MVS due to their 
more powerful representation capability in low-texture regions, reflections, \etc. 
Concretely, they infer the depth for each view from the 3D cost volume, 
which is constructed from the warped feature according to a set of predefined
depth hypotheses. Compared with hand-crafted 
similarity metrics in traditional methods, the 3D cost volume can 
capture more discriminative features to achieve more robust matching. Without 
loss of integrity, existing learning-based methods can be divided into two 
categories: {\em Regression} and {\em Classification}.

{\bf Regression} is the most primitive and straightforward implementation of 
the learning-based MVS method. 
It's a group of approaches \cite{yao2018mvsnet,luo2019p,cheng2020deep,
yang2020cost,gu2020cascade,yu2020fast} to regress the depth from the 3D cost volume 
through {\em Soft-argmin}, 
which softly weighting each depth hypothesis. More specifically, the model expects 
to regress greater weight for the depth hypothesis 
with a small cost. Theoretically, it can achieve the sub-pixel estimation of depth 
by weighted summation of discrete depth hypotheses. Nevertheless, the model 
needs to learn a complex combination of weights under indirect constraints performed on 
the weighted depth but not on the weight combination, which is non-trivial and 
tends to overfit. You can imagine 
that there are many weight combinations for a set of depth hypotheses that can be 
weighted and summed to the same depth, and this ambiguity also implicitly increases 
the difficulty of the model convergence.

{\bf Classification} is proposed in R-MVSNet 
\cite{yao2019recurrent} to infer 
the optimal depth hypothesis. Different from the weight estimation in 
regression, classification methods \cite{huang2018deepmvs,yao2019recurrent,yan2020dense} 
predict the probability of each depth hypothesis from the 3D cost volume 
and take the depth hypothesis with the maximum probability as the final 
estimation. Obviously, these methods cannot infer the exact depth 
directly from the model like regression methods. 
However, classification methods directly constrain the cost volume through the 
{\em cross-entropy} loss executed on the regularized probability volume, which is the 
essence of ensuring the robustness of MVS. 
Moreover, the estimated probability distribution can directly 
reflect the confidence, which is difficult to derive from the weight combination 
intuitively.
 
\begin{figure*}
  \begin{center}
     \includegraphics[trim={8.4cm 12.7cm 33.9cm 13cm},clip,width=1.0\linewidth]{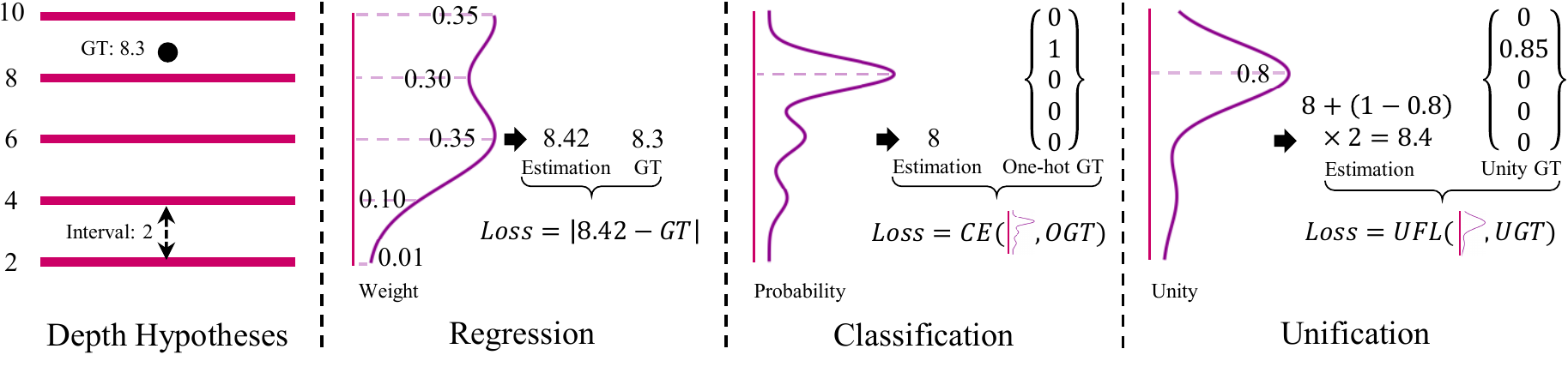}
  \end{center}
  \caption{{\bf Comparison with  different representations at a certain pixel.} The purple 
  curve represents different weights, probabilities, and unity of each depth hypothesis 
  obtained by regression, classification and unification respectively. 
  While the regression representation requires the exact weight of each hypothesis to 
  regress the depth, the classification representation only cares about which hypothesis 
  has the maximum probability and unification only needs to know the proximity with the 
  maximum unity. $CE$ denotes cross-entropy, and $UFL$ refers to Unified Focal Loss 
  (\cref{sec:ufl}).}
  \label{fig:unity}
\end{figure*}

In this paper, we seek to unify the advantages of regression and 
classification, that is, we hope that the model can accurately predict the 
depth while maintaining robustness. There is a fact 
that the depth hypothesis close to the ground-truth has more 
potential knowledge, while that of other remaining hypotheses is limited or 
even harmful due to the wrong induction of multimodal \cite{zhang2020adaptive}. 
Motivated by this, we present that estimating the weights for all depth 
hypotheses is redundant, and the model only needs to do regression on the 
{\em optimal depth hypothesis} that the representative depth interval (referring to the 
upper area until the next larger depth hypothesis) contains the ground truth depth. 
To achieve this, we propose a unified representation for depth, termed {\em Unification}. 
As shown in \cref{fig:unity}, unlike regression, the loss 
is executed on the regularized probability volume directly, and different from 
classification, our 
method estimates the {\em Unity} (What we call), whose label is composed by at most one non-zero 
continuous target $(0\sim1)$, to simultaneously represent the location of optimal 
depth hypothesis and its offset to the ground-truth depth. We take proximity 
(defined as the complement of the offset between ground-truth 
and optimal depth hypothesis) to characterize the 
non-zero target in unity label, which is more efficient than purely using offset. 
The detailed comparisons are in Supp. Mat. 

Moreover, we note that this unified representation faces an undeniable sample imbalance 
in both category and hardness. While Focal Loss (FL) \cite{lin2017focal} is 
the common solution proposed in the detection field, which is tailored to the 
traditional discrete label, the more general form (GFL) is proposed in 
\cite{gfl2020,zhang2021varifocalnet} to deal with the continuous label. Even though 
GFL has demonstrated its performance, we hold the belief that it has an obvious limitation 
in distinguishing hard and easy samples due to ignoring the magnitude of ground-truth. 
To this end, we put forward a more reasonable and unified form, 
called {\em Unified Focal Loss} (UFL), after thorough analysis to better address 
these challenges. In this way, the traditional FL can be regarded as a special case of 
UFL, while GFL is its imperfect expression. 

To demonstrate the superiority of our proposed modules, we present a coarse-to-fine 
framework termed {\bf UniMVSNet} (or {\bf UnifiedMVSNet}), named for 
its unification of depth representation and 
focal loss, which replaces the traditional representation of recent works 
\cite{cheng2020deep,yang2020cost,gu2020cascade} with {\em Unification} and adopts UFL for 
optimization. Extensive experiments show that our model surpasses all previous 
MVS methods and achieves state-of-the-art performance on both DTU 
\cite{aanaes2016large} and Tanks and Temples \cite{knapitsch2017tanks} benchmarks.

\section{Related Works}
\label{sec:related_works}

\noindent
{\bf Traditional MVS methods.} Taking the output scene representation as an axis of 
taxonomy, there are mainly four types of classic MVS methods: volumetric 
\cite{seitz1999photorealistic,kutulakos2000theory}, point cloud based 
\cite{lhuillier2005quasi,furukawa2009accurate}, mesh based \cite{fua1995object} and 
depth map based \cite{campbell2008using,schonberger2016pixelwise,schonberger2016structure,
xu2019multi,galliani2015massively}. Among them, the 
depth map based method is the most flexible one. Instead of operating in the 3D domain, 
it degenerates the complex 
problem of 3D geometry reconstruction to depth map estimation in the 2D domain. 
Moreover, as the intermediate representation, the estimated depth maps of all 
individual images can be merged into a consistent point cloud \cite{merrell2007real} 
or a volumetric reconstruction \cite{newcombe2011kinectfusion}, and the mesh can 
even be further reconstructed.

\noindent
{\bf Learning-based MVS methods.} While the traditional MVS pipeline mainly relies on 
hand-crafted similarity metrics, recent works apply deep learning for superior 
performance on MVS. SurfaceNet \cite{ji2017surfacenet} and LSM \cite{lsm2017} are the 
first proposed volumetric learning-based MVS pipelines to regress surface voxels from 
3D space. However, they are restricted to memory which is the common drawback of 
the volumetric representation. Most recently, MVSNet \cite{yao2018mvsnet} first realizes 
an end-to-end memory low-sensitive pipeline based on 3D cost volumes. This pipeline 
mainly consists of four steps: image feature extraction by 2D CNN, variance-based 
cost aggregation by homography warping, cost regularization through 3D CNN, and depth 
regression. To further excavate the potential capacity of this pipeline, some variants 
of MVSNet have been proposed, \eg, 
\cite{yao2019recurrent,gu2020cascade,cheng2020deep,yang2020cost} are proposed to reduce 
the memory requirement through RNN or coarse-to-fine manner, \cite{luo2019p,yi2020pyramid} 
are proposed to adaptively re-weight the contribution of different pixels in cost 
aggregation. Meanwhile, all existing methods are based on one of the two 
complementary of classification and regression to infer depth. In this paper, we propose 
a novel unified representation to integrate their advantages.

\begin{figure}[t]
  \begin{center}
     \includegraphics[trim={32cm 8.1cm 2cm 10cm},clip,width=1.0\linewidth]{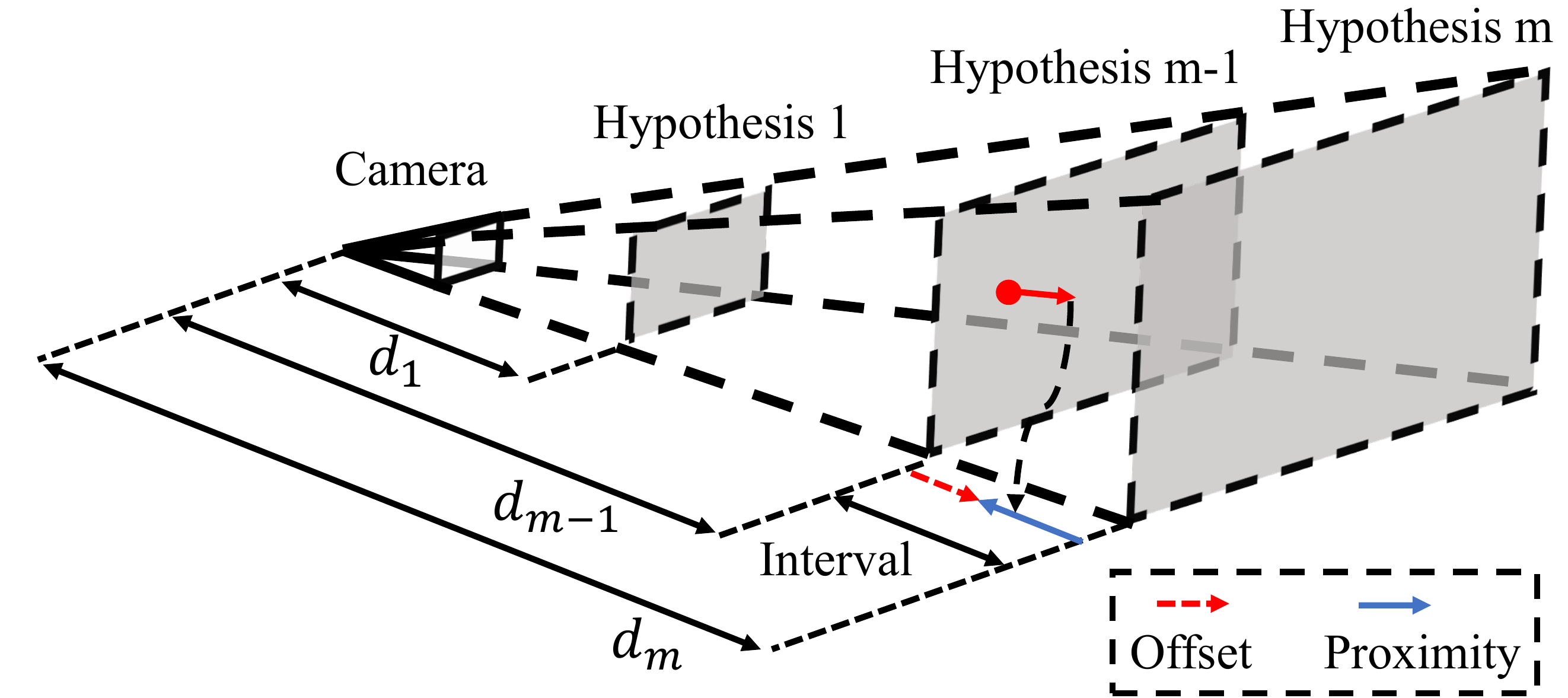}
  \end{center}
  \vspace{-0.4cm}
  \caption{{\bf Illustration of our {\em Unification}.} The 
  $(m-1)_{th}$ hypothesis is the optimal hypothesis for red point.}
  \label{fig:ill_unification}
  \vspace{-0.4cm}
\end{figure}

\section{Methodology}
\label{sec:methodology}
This section will present the main contributions of this paper in detail. 
We first review the common pipeline of the learning-based MVS approach in 
\cref{sec:review}, then introduce the proposed unified depth representation in 
\cref{sec:udr} and unified focal loss in \cref{sec:ufl}, and finally 
describe the detailed network architecture of our UniMVSNet in \cref{sec:unimvsnet}.

\subsection{Review of Learning-based MVS}
\label{sec:review}

Most end-to-end learning-based MVS methods are inherited from MVSNet 
\cite{yao2018mvsnet}, which constructs an elegant and effective pipeline 
to infer the depth $\mathbf{D}\in \mathbb{R}^{H' \times W'}$ of the reference 
image $\mathbf{I}_1$. Given 
multiple images $\{\mathbf{I}_i \in \mathbb{R}^{C \times H \times W}\}_{i=1}^N$ 
of a scene taken from $N$ different viewpoints, image features of all images 
$\{\mathbf{F}_i \in \mathbb{R}^{C' \times H' \times W'}\}_{i=1}^N$ are 
first extracted through a 2D network with shared weights. As mentioned above, 
the learning-based method is based on the 3D cost volume, and the depth 
hypothesis of $M$ layers $\{\mathbf{d}_i \in \mathbb{R}^{H' \times W'}\}_{i=1}^M$ 
is sampled from the whole known 
depth range to achieve this, where $\mathbf{d}_1$ represents the minimum depth 
and $\mathbf{d}_M$ represents the maximum depth. With this hypothesis, feature 
volumes $\{\mathbf{V}_i \in \mathbb{R}^{M \times C' \times H' \times W'}\}_{i=1}^N$ 
can be constructed in 3D space via differentiable homography by warping 2D 
image features of source images to the reference camera frustum. The homography 
between the feature maps of $i_{th}$ view and the reference feature maps at 
depth $d$ is expressed as:
\begin{equation}
  \mathbf{H}_i(d) = d \mathbf{K}_i \mathbf{T}_i \mathbf{T}_1^{-1} \mathbf{K}_1^{-1}
\label{eq:variance}
\end{equation}
where $\mathbf{K}$ and $\mathbf{T}$ refer to camera intrinsics and 
extrinsics respectively.

To handle arbitrary number of input views, 
the multiple feature volumes $\{ V_i \}_{i=1}^N$ need to be aggregated to 
one cost volume $\mathbf{C} \in \mathbb{R}^{M \times C' \times H' \times W'}$. 
The aggregation strategy consists two dominant groups: statistical and 
adaptive. The variance-based mapping is a typical statistical aggregation: 
\begin{equation}
  \mathbf{C} = \frac{1}{N}\sum_{i=1}^N (\mathbf{V}_i - \mathbf{\bar{V}})^2
\label{eq:homography}
\end{equation}
Where $\mathbf{\bar{V}}$ denotes the average feature volume. Furthermore, the 
adaptive aggregation is proposed to re-weight the contribution of different pixels, 
which can be modeled as:
\begin{equation}
  \mathbf{C} = \frac{1}{N-1} \sum_{i=2}^N \mathcal{W}_i \odot (\mathbf{V}_i - \mathbf{V}_1)^2
\label{eq:adaptation}
\end{equation}
where $\mathcal{W}$ is the learnable weight generated by an auxiliary network, 
and $\odot$ denotes element-wise multiplication. 

The matching cost between the reference view and all source views under each depth 
hypothesis has been encoded into the cost volume, which is required 
to be further refineed to generate a probability volume 
$\mathbf{P} \in \mathbb{R}^{M \times H' \times W'}$ through a softmax-based 
regularization network. Concretely, the probability volume is treated as the 
weight of depth hypotheses in regression methods and the depth at pixel $(x,y)$ is 
calculated as the sum of the weighted hypotheses as:
\begin{equation}
  \mathbf{D}^{x,y} = \sum_{d=\mathbf{d}_1^{x,y}}^{\mathbf{d}_M^{x,y}} d \mathbf{P}(d)^{x,y}
\label{eq:regression}
\end{equation}
and the model is constrained by the {\em L1} loss between $\mathbf{D}$ and the 
ground-truth depth. In classification methods, $\mathbf{P}$ refers to the probability 
of depth hypotheses and the depth is estimated as the hypothesis whose 
probability is maximum: 
\begin{equation}
  \mathbf{D}^{x,y} = \mathop{\arg\max}_{d \in \{\mathbf{d}_i^{x,y}\}_{i=1}^M} \mathbf{P}(d)^{x,y}
\label{eq:regression}
\end{equation}
and the model is trained by the {\em cross-entropy} loss between $\mathbf{P}$ and the 
ground-truth one-hot probability volume.

In traditional one-stage methods, compared with original input images, the depth 
map is either downsized during feature extraction \cite{yao2018mvsnet} or before 
the input \cite{yi2020pyramid} to save memory, while in the coarse-to-fine method 
\cite{gu2020cascade, yang2020cost, cheng2020deep}, 
it's a multi-scale result $\{\mathbf{D}_i\}_{i=1}^L$ with incremental resolution 
generated by repeating the above pipeline $L$ times. The multi-scale is realized 
by a FPN-like \cite{lin2017feature} feature extraction network, and the depth 
hypothesis with decreasing depth range is sampled based on the depth map generated 
in the previous stage.

\SetKwInOut{KwIni}{Initialization}
\SetKw{KwAnd}{and}
\begin{algorithm}[t]
  \caption{Unity Generation}
  \label{ag:ug}
  \LinesNumbered
  \KwIn{Ground-truth depth $\mathbf{D}_{gt} \in \mathbb{R}^{H' \times W'}$; 
  Depth hypotheses $\{\mathbf{d}_i \in \mathbb{R}^{H' \times W'}\}_{i=1}^M$.}
  \KwOut{Ground-truth Unity $\{\mathbf{U}_i \in \mathbb{R}^{H' \times W'}\}_{i=1}^M$.}
  \KwIni{Depth interval $r=0$.}
  \For{$i=1$ \KwTo $M$}{
    \For{$(x,y)=(1,1)$ \KwTo $(H',W')$}{
      \If{$i<M$}{
        $r=\mathbf{d}_{i+1}^{x,y}-\mathbf{d}_i^{x,y}$ \;
      }
      \eIf{$\mathbf{d}_i^{x,y}\le\mathbf{D}_{gt}^{x,y}$ \KwAnd $\mathbf{d}_i^{x,y}+r>\mathbf{D}_{gt}^{x,y}$}{
        $\mathbf{U}_i^{x,y}=1 - \frac{\mathbf{D}_{gt}^{x,y} - \mathbf{d}_i^{x,y}}{r}$ \;
      }{
        $\mathbf{U}_i^{x,y}=0$ \;
      }
    }
  }
  \KwRet{$\{\mathbf{U}_i\}_{i=1}^M$}.
\end{algorithm} 

\begin{algorithm}[t]
  \caption{Unity Regression}
  \label{ag:ur}
  \LinesNumbered
  \KwIn{Estimated unity $\{\mathbf{\widehat{U}}_i \in \mathbb{R}^{H' \times W'}\}_{i=1}^M $; 
  Depth hypotheses $\{\mathbf{d}_i \in \mathbb{R}^{H' \times W'}\}_{i=1}^M$.}
  \KwOut{Regressed Depth $\mathbf{D} \in \mathbb{R}^{H' \times W'}$.}
  \KwIni{Depth interval $r=0$.}
  \For{$(x,y)=(1,1)$ \KwTo $(H',W')$}{
    Optimal hypothesis index $o=\mathop{\arg\max}\limits_{i \in \{1,\cdots,M\}} \mathbf{\widehat{U}}_i^{x,y}$\;
    Optimal hypothesis $d=\mathbf{d}_o^{x,y}$ \;
    \eIf{$o<M$}{
      $r=\mathbf{d}_{o+1}^{x,y}-\mathbf{d}_o^{x,y}$ \;
    }{
      // previous interval for the last hypothesis \

      $r=\mathbf{d}_{o}^{x,y}-\mathbf{d}_{o-1}^{x,y}$ \;
    }
    Offset $off=(1-\mathbf{\widehat{U}}_o^{x,y}) \times r$ \;
    Depth $\mathbf{D}^{x,y}=d+off$ \;
  }
  \KwRet{$\mathbf{D}$}.
\end{algorithm} 

\begin{figure*}
  \begin{center}
     \includegraphics[trim={13.8cm 7.9cm 12.8cm 14.5cm},clip,width=1.0\linewidth]{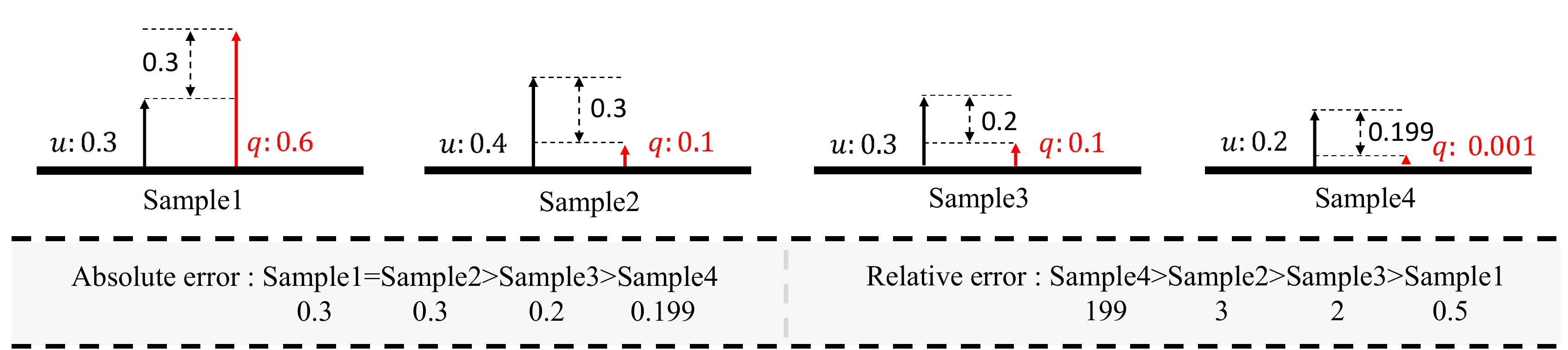}
  \end{center}
  \vspace{-0.4cm}
  \caption{{\bf Hardness measurement of different samples.} Four typical samples with 
  different ground-truth $q$ and different estimation $u$.}
  \label{fig:ufl_sample}
  \vspace{-0.4cm}
\end{figure*}

\subsection{Unified Depth Representation}
\label{sec:udr}

As aforementioned, the regression method tends to overfit due to its indirect learning 
cost volume and ambiguity in the correspondence 
between depth and the weight combination. For the classification method, although 
it can constrain the cost volume directly, it cannot predict exact 
depth like regression methods due to its discrete prediction. In this paper, we found 
that they can complement each other, and we unify them successfully through 
our unified depth representation, as shown in \cref{fig:ill_unification}. We recast the 
depth estimation as a multi-label 
classification task, in which the model needs to classify which hypothesis is 
the optimal one and regress the proximity for it. In other words, we first 
adopt classification to narrow the depth range of the final regression, but they are 
executed simultaneously in our implementation. Therefore, 
the model in our {\em Unification} representation is able to estimate an accurate depth 
like regression 
methods, and it also directly optimizes the cost volume like classification methods. 
Below, we will introduce how to generate the ground-truth unity from ground-truth 
depth (Unity generalization), and how to regress the depth from the estimated unity 
(Unity regression).

{\bf Unity generation:} As shown in \cref{fig:unity}, ground-truth unity 
$\{\mathbf{U}_i\}_{i=1}^M$ is a more general form of one-hot label peaked at the 
optimal depth hypothesis whose depth interval contains ground-truth depth. The 
at most one non-zero target is a continuous number and represents the proximity of the 
optimal hypothesis to ground-truth depth. The detail of unity generation is shown 
in \cref{ag:ug}, which is one more step of proximity calculation than the 
one-hot label generation in classification methods.

{\bf Unity regression:} Unlike the traditional way of predicting the probability 
volume by {\em softmax} operators, unification representation estimated it 
through {\em sigmoid} operators. 
Here, we disassemble the estimated probability volume $\mathbf{P}$ into the estimated 
unity $\{\mathbf{\widehat{U}}_i\}_{i=1}^M$ along the $M$ dimension.
To regress the depth, we first select the 
optimal hypothesis with the maximum unity at each pixel, then calculate the offset to 
ground-truth depth, and finally fuse the estimated depth. The detailed procedure 
is shown in \cref{ag:ur}. 

\begin{figure*}
  \begin{center}
     \includegraphics[trim={0.3cm 5.8cm 8.3cm 0.5cm},clip,width=1.0\linewidth]{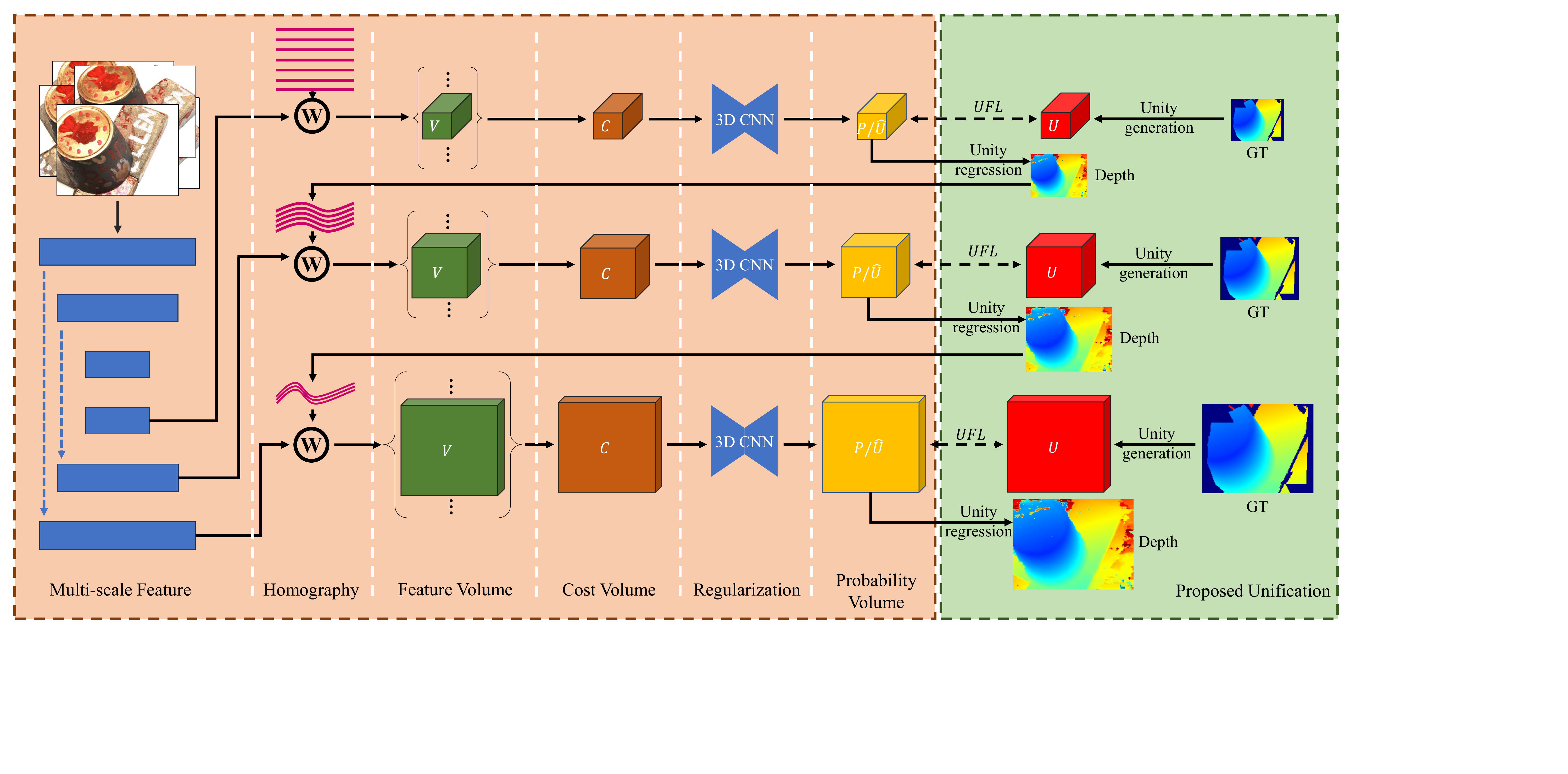}
  \end{center}
  \vspace{-0.4cm}
  \caption{{\bf Illustration of UniMVSNet.} This is a typical coarse-to-fine framework. The 
  part with pink background is inherited from existing methods, and the green part is our 
  novel modules. The depth hypothesis is represented by the red curve for convenience.} 
  \label{fig:framework}
  \vspace{-0.4cm}
\end{figure*}

\subsection{Unified Focal Loss}
\label{sec:ufl}
Generally, the depth hypothesis of MVS models will be sampled quite densely to ensure 
the accuracy of the estimated depth, which will cause obvious sample imbalance due to 
only one positive sample (the at most one non-zero target) among hundreds of 
hypotheses. Meanwhile, the model needs 
to pay more attention to hard samples to prevent overfitting. Relevant Focal Loss (FL) 
\cite{lin2017focal} has been proposed to solve these two problems, which automatically 
distinguishes hard samples through the estimated unity $u \in [0,1]$ and rebalances 
the sample through tunable parameter $\alpha$ and $\gamma$. Here, we discuss a 
certain pixel for convenience. The typical definition of FL is:
\begin{equation}
  {\rm FL}(u,q)=
  \begin{cases} 
    -\alpha(1-u)^\gamma\log(u),  & q=1 \\
    -(1-\alpha)u^\gamma\log(1-u), & \text{else}
  \end{cases}
\label{eq:fl}
\end{equation}
where $q \in \{0,1\}$ is the discrete target. Therefore, the traditional FL is not 
suitable for our continuous situation. To enable successful training under the case 
of our representation, we borrow the main idea from FL. 
Above all, the binary cross-entropy $-\log(u)$ or $-\log(1-u)$ needs to be extended 
to its complete form ${\rm BCE}(u, q)=-q\log(u)-(1-q)\log(1-u)$. Correspondingly, the 
scaling factor should also be adjusted appropriately. The generalized FL form (GFL) 
obtained through these two steps is:
\begin{equation}
  {\rm GFL}(u,q)=
  \begin{cases} 
    \alpha|q-u|^\gamma {\rm BCE}(u, q),  & q>0 \\
    (1-\alpha)u^\gamma {\rm BCE}(u, q), & \text{else}
  \end{cases}
\label{eq:gfl}
\end{equation}
where $q \in [0,1]$ is the continuous target. This advanced version is currently 
adopted by some existing methods with different 
names, \eg, QFL in \cite{gfl2020} or VFL in \cite{zhang2021varifocalnet}. But in this 
paper, we point out that this implementation is not perfect in scaling 
hard and easy samples, because they ignore the magnitude of the ground-truth.

As shown in \cref{fig:ufl_sample}, the first two samples will be considered 
the hardest under the absolute error $|q-u|$ measurement in GFL. However, the 
absolute error cannot distinguish samples with different targets. Even if the 
first two samples in \cref{fig:ufl_sample} have the same absolute error, this 
error obviously has a smaller effect on the first sample due to its larger ground-truth. 
To solve this ambiguity, we further improve the scaling factor in GFL through 
relative error and propose our naive version of Unified Focal Loss (UFL) just as:
\begin{equation}
  {\rm UFL}(u,q)=
  \begin{cases} 
    \alpha(\frac{|q-u|}{q^+})^\gamma {\rm BCE}(u, q), \!  & q>0 \\
    (1-\alpha)(\frac{u}{q^+})^\gamma {\rm BCE}(u, q), \! & \text{else}
  \end{cases}
\label{eq:ufl_naive}
\end{equation}
where $q^+ \in (0,1]$ is the positive target. It can be seen 
from \cref{eq:ufl_naive} that FL is a special case of UFL when the positive 
target is the constant 1.

Moreover, we noticed that the range of scaling factor $\frac{|q-u|}{q^+}$ is 
$[0, +\infty)$, which may lead to a special case like the last sample in 
\cref{fig:ufl_sample}. Even a small number of such samples will overwhelm the loss 
and computed gradients due to their huge scaling factor. In this paper, we solve 
this problem by introducing a dedicated function to control the range of the scaling 
factor. 
Meanwhile, to keep the precious positive learning signals, we adopt an asymmetrical 
scaling strategy. And the complete UFL can be modeled as:
\begin{equation}
  {\rm UFL}(u,q)\!=\!
  \begin{cases} 
    \alpha^+ (S_b^+(\frac{|q-u|}{q^+}))^\gamma {\rm BCE}(u, q), \!  & q>0 \\
    \alpha^- (S_b^-(\frac{u}{q^+}))^\gamma {\rm BCE}(u, q), \! & \text{else}
  \end{cases}
\label{eq:ufl}
\end{equation}
where the dedicated function $S_b(x)$ is designed as the sigmoid-like function 
($1 / (1 + b^{-x})$) with a base of $b$ in this paper.

\subsection{UniMVSNet}
\label{sec:unimvsnet}
It's straightforward to apply our Unification and UFL to existing learning-based 
MVS methods. To illustrate the effectiveness and flexibility of the proposed 
modules, we build the UniMVSNet, whose framework is depicted in 
\cref{fig:framework}, based on the coarse-to-fine strategy. This pipeline abides 
by the procedure reviewed in \cref{sec:review}, except the depth 
representation and optimization.

Inherited from CasMVSNet \cite{gu2020cascade}, We adopt a FPN-like network 
to extract multi-scale features, and uniformly sample the depth hypothesis with 
a decreasing interval and a decreasing number. To better handle the unreliable 
matching in non-Lambertian regions, we adopt an adaptive aggregation method 
with negligible parameters increasing like \cite{yi2020pyramid} to aggregate 
the feature volumes warped by the differentiable homography. Meanwhile, we also 
apply multi-scale 3D CNNs to regularize the cost volume, and the generated 
probability volume $\mathbf{P}$ at each stage is treated as the estimated Unity 
$\{\mathbf{\widehat{U}}_i\}_{i=1}^M$ here, which can be further regressed to accurate 
depth as shown in \cref{ag:ur}. It can be seen from \cref{fig:framework} 
that UniMVSNet directly optimizes cost volume through UFL, which can effectively 
avoid the overfitting of indirect learning strategy in regression methods.

{\bf Training Loss. } As shown in \cref{fig:framework}, we apply UFL to all 
stages and fuse them with different weights. The total loss can be defined as:
\begin{equation}
  Loss = \sum_{i=1}^L \lambda_i \overline{{\rm UFL}}_i
\label{eq:loss}
\end{equation}
where $\overline{{\rm UFL}}_i$ is the average of UFL of all valid pixels at stage $i$ and 
$\lambda_i$ denotes the weight of $\overline{{\rm UFL}}$ at $i_{th}$ stage.

\section{Experiments}
\label{sec:experiments}

\begin{figure}[t]
  \begin{center}
     \includegraphics[trim={7cm 2cm 7cm 2cm},clip,width=1.0\linewidth]{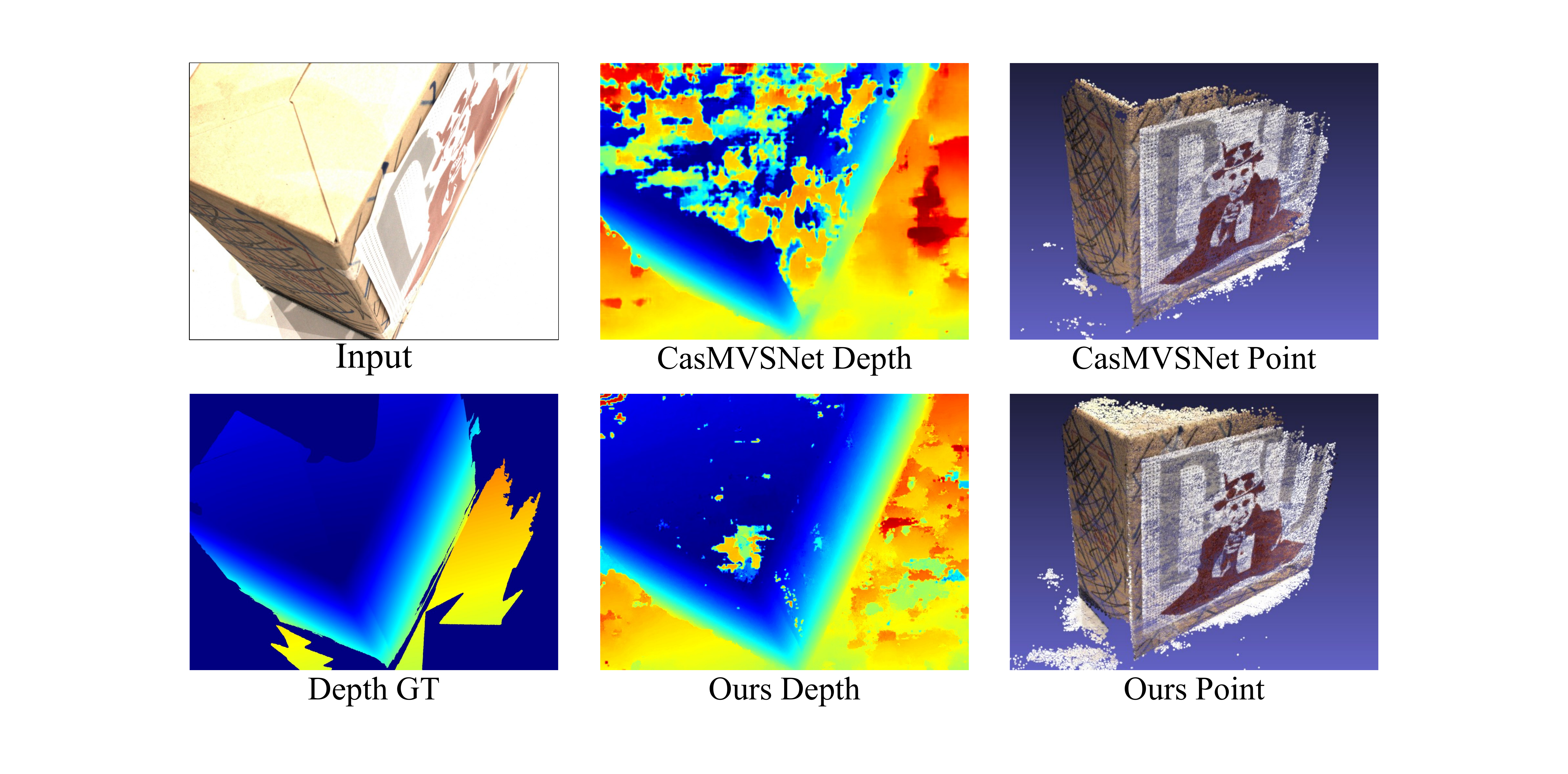}
  \end{center}
  \vspace{-0.4cm}
  \caption{{\bf Depth estimation and point reconstruction of scan 13 on DTU.} Our model 
  produces more accurate and complete results.}
  \label{fig:dtu_depth}
  \vspace{-0.4cm}
\end{figure}

\begin{table}
  \begin{center}
  \resizebox{1.0\linewidth}{!}{
  \begin{tabular}{l|ccc}
  \toprule
  Method & ACC.(mm) & Comp.(mm) & Overall(mm) \\
  \hline
  Furu \cite{furukawa2009accurate} & 0.613 & 0.941 & 0.777 \\
  Gipuma \cite{galliani2015massively} & {\bf 0.283} & 0.873 & 0.578 \\
  COLMAP \cite{schonberger2016structure,schonberger2016pixelwise} & 0.400 & 0.664 & 0.532 \\
  SurfaceNet \cite{ji2017surfacenet} & 0.450 & 1.040 & 0.745 \\
  MVSNet \cite{yao2018mvsnet} & 0.396 & 0.527 & 0.462 \\
  P-MVSNet \cite{luo2019p} & 0.406 & 0.434 & 0.420 \\
  R-MVSNet \cite{yao2019recurrent} & 0.383 & 0.452 & 0.417 \\
  Point-MVSNet \cite{chen2019point} & 0.342 & 0.411 & 0.376 \\
  AA-RMVSNet \cite{wei2021aa} & 0.376 & 0.339 & 0.357 \\
  CasMVSNet \cite{gu2020cascade} & 0.325 & 0.385 & 0.355 \\
  CVP-MVSNet \cite{yang2020cost} & 0.296 & 0.406 & 0.351 \\
  UCS-Net \cite{cheng2020deep} & 0.338 & 0.349 & 0.344 \\
  \hline
  UniMVSNet (ours) & 0.352 & {\bf 0.278} & {\bf 0.315} \\
  \bottomrule
  \end{tabular}
  }
  \end{center}
  \vspace{-0.4cm}
  \caption{{\bf Quantitative results on DTU evaluation set.} Best results in 
  each category are in {\bf bold}. Our model ranks first in terms of Completeness 
  and Overall metrics.}
  \label{tb:dtu_compare}
  \vspace{-0.6cm}
\end{table}

\begin{figure*}
  \begin{center}
     \includegraphics[trim={2.6cm 8.8cm 2.6cm 9.7cm},clip,width=1.0\linewidth]{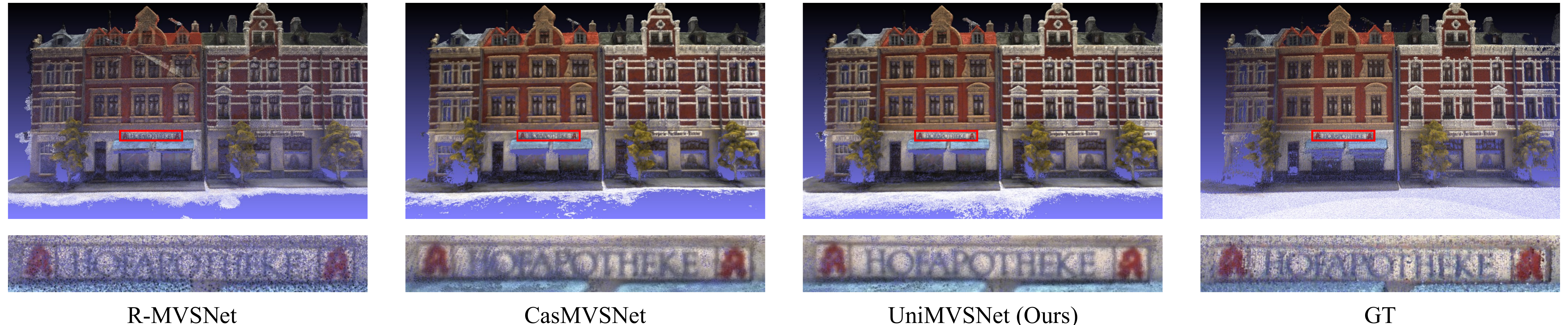}
  \end{center}
  \vspace{-0.3cm}
  \caption{{\bf Qualitative results of scan 15 on DTU.} The top row shows the point clouds 
  generated by different methods and the ground-truth, and the bottom row shows a more 
  detailed local region corresponding to the red rectangle.}
  \label{fig:dtu_point}
\end{figure*}

\begin{table*}
  \begin{center}
  \resizebox{1.0\linewidth}{!}{
  \begin{tabular}{l|ccccccccc|ccccccc}
  \toprule
  \multirow{2}*{Method} & \multicolumn{9}{c|}{Intermediate} & \multicolumn{7}{c}{Advanced} \\
  \cline{2-17}
  & Mean & Fam. & Fra. & Hor. & Lig. & M60 & Pan. & Pla. & Tra. & Mean & Aud. & Bal. & Cou. & Mus. & Pal. & Tem. \\
  \hline
  Point-MVSNet \cite{chen2019point} & 48.27 & 61.79 & 41.15 & 34.20 & 50.79 & 51.97 & 50.85 & 52.38 & 43.06 & - & - & - & - & - & - & - \\
  PatchmatchNet \cite{wang2021patchmatchnet} & 53.15 & 66.99 & 52.64 & 43.24 & 54.87 & 52.87 & 49.54 & 54.21 & 50.81 & 32.31 & 23.69 & 37.73 & 30.04 & 41.80 & 28.31 & 32.29 \\
  UCS-Net \cite{cheng2020deep} & 54.83 & 76.09 & 53.16 & 43.03 & 54.00 & 55.60 & 51.49 & 57.38 & 47.89 & - & - & - & - & - & - & - \\
  CVP-MVSNet \cite{yang2020cost} & 54.03 & 76.50 & 47.74 & 36.34 & 55.12 & 57.28 & 54.28 & 57.43 & 47.54 & - & - & - & - & - & - & - \\
  P-MVSNet \cite{luo2019p} & 55.62 & 70.04 & 44.64 & 40.22 & {\bf 65.20} & 55.08 & 55.17 & 60.37 & 54.29 & - & - & - & - & - & - & - \\
  CasMVSNet \cite{gu2020cascade} & 56.84 & 76.37 & 58.45 & 46.26 & 55.81 & 56.11 & 54.06 & 58.18 & 49.51 & 31.12 & 19.81 & 38.46 & 29.10 & 43.87 & 27.36 & 28.11 \\
  ACMP \cite{xu2020planar} & 58.41 & 70.30 & 54.06 & {\bf 54.11} & 61.65 & 54.16 & 57.60 & 58.12 & 57.25 & 37.44 & 30.12 & 34.68 & {\bf 44.58} & 50.64 & 27.20 & 37.43 \\
  $D^2$HC-RMVSNet \cite{yan2020dense} & 59.20 & 74.69 & 56.04 & 49.42 & 60.08 & 59.81 & 59.61 & 60.04 & 53.92 & - & - & - & - & - & - & - \\
  VisMVSNet \cite{zhang2020visibility} & 60.03 & 77.40 & 60.23 & 47.07 & 63.44 & 62.21 & 57.28 & 60.54 & 52.07 & 33.78 & 20.79 & 38.77 & 32.45 & 44.20 & 28.73 & {\bf 37.70} \\
  AA-RMVSNet \cite{wei2021aa} & 61.51 & 77.77 & 59.53 & 51.53 & 64.02 & 64.05 & 59.47 & 60.85 & 55.50 & 33.53 & 20.96 & 40.15 & 32.05 & 46.01 & 29.28 & 32.71 \\
  EPP-MVSNet \cite{ma2021epp} & 61.68 & 77.86 & 60.54 & 52.96 & 62.33 & 61.69 & 60.34 & 62.44 & 55.30 & 35.72 & 21.28 & 39.74 & 35.34 & 49.21 & 30.00 & 38.75 \\
  \hline
  UniMVSNet (ours) & {\bf 64.36} & {\bf 81.20} & {\bf 66.43} & 53.11 & 63.46 & {\bf 66.09} & {\bf 64.84} & {\bf 62.23} & {\bf 57.53} & {\bf 38.96} & {\bf 28.33} & {\bf 44.36} & 39.74 & {\bf 52.89} & {\bf 33.80} & 34.63 \\
  \bottomrule
  \end{tabular}
  }
  \end{center}
  \vspace{-0.4cm}
  \caption{{\bf Quantitative results of F-score on Tanks and Temples benchmark.} 
  Best results in each category are in {\bf bold}. ``Mean'' refers to the mean F-score of 
  all scenes (higher is better). 
  Our model outperforms all previous MVS methods with a significant margin on both 
  Intermediate and Advanced set.}
  \label{tb:tank_compare}
  \vspace{-0.4cm}
\end{table*}

This section demonstrates the start-of-the-art performance of UniMVSNet with 
comprehensive experiments and verifies the effectiveness of the proposed Unification 
and UFL through ablation studies. We first introduce the datasets and 
implementation and then analyze our results.

\noindent
{\bf Datasets.} We evaluate our model on DTU \cite{aanaes2016large} and 
Tanks and Temples \cite{knapitsch2017tanks} benchmark and finetune on 
BlendedMVS \cite{yao2020blendedmvs}. (a) DTU is an 
indoor MVS dataset with 124 different scenes scanned from 49 or 64 views 
under 7 different lighting conditions with fixed camera trajectories. We adopt 
the same training, validation, and evaluation split as defined in 
\cite{yao2018mvsnet}. (b) Tanks and Temples is collected in a more complex 
realistic environment, and it's divided into the intermediate and advanced set. 
While intermediate set contains 8 scenes with large-scale variations, advanced 
set has 6 scenes. (c) BlendedMVS is a 
large-scale synthetic dataset, which is consisted of 113 indoor and outdoor 
scenes and is split into 106 training scenes and 7 validation scenes.

\noindent
{\bf Implementation.} Following the common practice, we first train our model 
on the DTU training set and evaluate on DTU evaluation set, and then finetune our 
model on BlendedMVS before 
validating the generalization of our approach on Tanks and Temples. The 
input view selection and data pre-processing strategies are the same as 
\cite{yao2018mvsnet}. Meanwhile, we utilize the finer DTU ground-truth as 
\cite{wei2021aa}. In this paper, UniMVSNet is implemented in 3 stages 
with 1/4, 1/2 ,and 1 of original input image resolution respectively. We 
follow the same configuration (\eg, depth interval) of the model at 
each stage as \cite{gu2020cascade} 
in both training and evaluation of DTU. When training on DTU, the number of input 
images is set to $N=5$ and the image resolution is resized to $640 \times 512$. 
To emphasize the contribution of positive signals, we set $\alpha^+=1$, and scale 
the range of $S_5^+$ to $[1,3)$ and $S_5^-$ to $[0, 1)$.
The other tunable parameters in UFL are configured stage-wise, \eg, 
$\alpha^-$ is set to 0.75, 0.5, and 0.25, and $\gamma$ is set to 2, 1, and 0 from the 
coarsest stage to the finest stage.. We optimize our model for 16 epochs 
with Adam optimizer \cite{kingma2014adam}, and the initial learning rate is 
set to 0.001 and decayed by 2 after 10, 12, and 14 epochs. During the evaluation of 
DTU, we also resize the input image size to $1152 \times 864$ and set the number 
of the input images to 5. We report the standard metrics (accuracy, 
completeness, and overall) proposed by the official evaluation protocol 
\cite{aanaes2016large}. Before testing on Tanks and Temples benchmark, we finetune 
our model on BlendedMVS for 10 epochs. We take 7 images 
as the input with the original size of $768 \times 576$. For benchmarking on 
Tanks and Temples, the number of depth hypotheses in the coarsest stage is changed 
from 48 to 64, and the corresponding depth interval is set to 3 times as the 
interval of \cite{yao2018mvsnet}. We set the number of input images to 
11 and report the F-score metric

\subsection{Results on DTU}

Similar to previous methods \cite{yao2018mvsnet,yao2019recurrent,gu2020cascade}, we 
introduce photometric and geometric constraints for depth map filtering. The 
probability threshold and the number of consistent views are set to 
0.3 and 3 respectively, which is the same as \cite{yao2019recurrent}. The final 3D 
point cloud is obtained through the same depth map fusion method as 
\cite{yao2018mvsnet,yao2019recurrent,gu2020cascade}.

We compare our method to those traditional and recent learning-based MVS methods. The 
quantitative results on the DTU evaluation set are summarized in \cref{tb:dtu_compare}, 
which indicates that our method has made great progress in performance. While Gipuma 
\cite{galliani2015massively} ranks first in the accuracy metric, our method outperforms 
all methods on the other two metrics significantly. Depth map estimation and point 
reconstruction of a reflective and low-textured sample are shown in 
\cref{fig:dtu_depth}, which shows that our model is more robust on the challenge 
regions. \Cref{fig:dtu_point} shows some qualitative results compared with 
other methods. We can see that our model can generate more complete point 
clouds with finer details.

\begin{figure*}
  \begin{center}
     \includegraphics[trim={2cm 8.2cm 2cm 9.9cm},clip,width=1.0\linewidth]{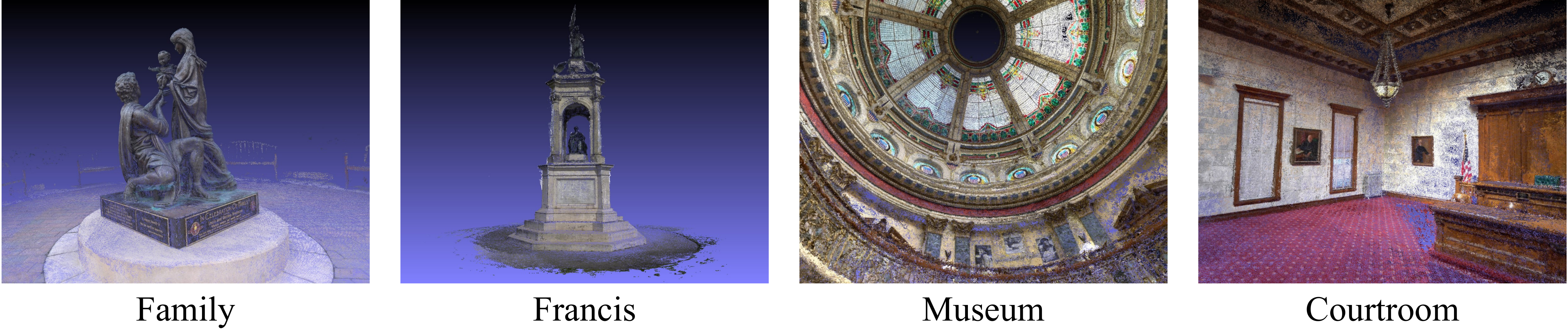}
  \end{center}
  \vspace{-0.5cm}
  \caption{{\bf Qualitative results of some scenes on Tanks and Temples.}}
  \label{fig:tank_point}
  \vspace{-0.2cm}
\end{figure*}

\begin{table*}
  \begin{center}
  \resizebox{1.0\linewidth}{!}{
  \begin{tabular}{l|ccc|ccccc|cc|c|c|ccc}
  \toprule
  \multirow{2}*{Method} & \multicolumn{3}{c|}{Representation} & \multicolumn{5}{c|}{Loss Function} & \multicolumn{2}{c|}{Aggregation} & \multirow{2}*{FGT} & \multirow{2}*{Input} & \multirow{2}*{ACC.(mm)} & \multirow{2}*{Comp.(mm)} & \multirow{2}*{Overall.(mm)} \\
  \cline{2-11}
  & Reg & Cla & Uni & L1 & CE & BCE & GFL & UFL & Adaptive & Variance & & & & &  \\
  \hline
  Baseline (Reg) & \checkmark & & & \checkmark & & & & & & \checkmark & & 3 & 0.369 & 0.317 & 0.343 \\
  Baseline (Reg) & \checkmark & & & \checkmark & & & & & & \checkmark & & 5 & 0.368 & 0.312 & 0.340 \\
  \hline
  Baseline (Cla) & & \checkmark & & & \checkmark & & & & & \checkmark & & 5 & 0.425 & 0.285 & 0.355 \\
  \hline
  Baseline (Uni) & & & \checkmark & & & \checkmark & & & & \checkmark & & 5 & 0.372 & 0.282 & 0.327 \\
  Baseline (Uni) + GFL & & & \checkmark & & & & \checkmark & & & \checkmark & & 5 & 0.361 & 0.289 & 0.325 \\
  Baseline (Uni) + UFL & & & \checkmark & & & & & \checkmark & & \checkmark & & 5 & 0.353 & 0.287 & 0.320 \\
  Baseline (Uni) + UFL + AA & & & \checkmark & & & & & \checkmark & \checkmark & & & 5 & 0.355 & 0.279 & 0.317 \\
  Baseline (Uni) + UFL + AA + FGT & & & \checkmark & & & & & \checkmark & \checkmark & & \checkmark & 5 & {\bf 0.352} & {\bf 0.278} & {\bf 0.315} \\
  \bottomrule
  \end{tabular}
  }
  \end{center}
  \vspace{-0.5cm}
  \caption{{\bf Ablation results on DTU evaluation set.} ``AA'' 
  and ``FGT'' refer to adaptive aggregation and finer 
  ground-truth respectively. 
  ``Baseline (Reg)'' is the original CasMVSNet \cite{gu2020cascade}. We set the confidence 
  threshold and the consistent views to 0.3 and 3 for all models. }
  \label{tb:ab}
  \vspace{-0.4cm}
\end{table*}

\subsection{Results on Tanks and Temples}

As the common practice, we verify the generalization ability of our method on Tanks 
and Temples benchmark using the model 
finetuned on BlendedMVS. We 
adopt a depth map filtering strategy similar to that of DTU, except for the geometric 
constraint. Here, we follow the dynamic geometric consistency checking strategies 
proposed in \cite{yan2020dense}. Through this dynamic method, those pixels with fewer 
consistent views but smaller reprojection errors and those with larger errors but more 
consistent views will also survive.

The corresponding quantitative results on both intermediate and advanced sets are reported 
in \cref{tb:tank_compare}. 
Our method achieves state-of-the-art performance among all existing MVS
methods and yields first place in most scenes. Notably, our model outperforms the previous 
best model by 2.68 points and 3.24 points on the intermediate and advanced sets. Such 
obvious advantages just show that 
our model not only has the best performance but also exhibits the strongest 
generalization and robustness. The qualitative point cloud results are visualized in 
\cref{fig:tank_point}.

\subsection{Ablation Studies}
\label{sec:as}
As aforementioned, we adopt some extra strategies (\eg, adaptive aggregation and 
finer ground-truth) that have been adopted by recent methods \cite{wei2021aa,yi2020pyramid} 
to train our model for a fair comparison with them. However, this may not be fair to 
those methods inherited only from MVSNet.
In this section, we will prove through extensive ablation studies that even if these 
strategies are eliminated, our method still has a significant improvement. We use our 
baseline CasMVSNet \cite{gu2020cascade}, whose original representation is regression, 
as the backbone and changing various components, \eg, depth representation, optimization, 
aggregation, and ground-truth. And we adopt 5 input views for all models for a fair 
comparison.

\noindent
{\bf Benefits of Unification.} As shown in \cref{tb:ab}, significant progress can 
be made even if purely replacing the traditional depth representation with our 
unification. Meanwhile, unification is more robust when the hypothesis range of the 
finer stage doesn't cover the ground-truth depth. In this case, the target unity of 
the unification representation generated by \cref{ag:ug} is all zero, which is 
a correct supervision signal anyway, and the traditional representation will generate 
an incorrect supervision signal to pollute the model training. Meanwhile, our Unification 
can also generate sharp depth on object boundaries like \cite{tosi2021smd}.

\noindent
{\bf Benefits of UFL.} Applying Focal Loss to our representation can effectively 
overcome the sample imbalance problem. It can be seen from \cref{tb:ab} that GFL 
has a huge benefit to accuracy, albeit with a slight loss of completeness. 
And our UFL can further improve the accuracy and completeness significantly on the basis 
of GFL. 
More ablation results about UFL are shown in Supp. Mat. 

\noindent
{\bf Other strategies and less data.} \cref{tb:ab_other_less} shows that our Unification performs 
better and is more concise compared to other strategies. Meanwhile, we still achieve 
excellent performance even with only 50\% of the training data.

\begin{table}[t]
  \begin{center}
  \resizebox{1.0\linewidth}{!}{
  \begin{tabular}{l|cc|ccc}
  \hline
  Method & Confidence & Consistent View & ACC.(mm) & Comp.(mm) & Overall(mm) \\
  \hline
  Baseline (Cla) + Regress finetune & 0.3 & 3 & 0.371 & 0.295 & 0.333 \\
  \hline
  Baseline (Uni) + UFL (50\% data) & 0.3 & 3 & 0.364 & 0.284 & 0.324 \\
  \hline
  \end{tabular}
  }
  \end{center}
  \vspace{-0.4cm}
  \caption{{\bf Some other Ablation results.}}
  \label{tb:ab_other_less}
  \vspace{-0.5cm}
\end{table}



\section{Conclusion}
\label{sec:conclusion}

In this paper, we propose a unified depth representation and a unified focal loss to 
promote the effectiveness of multi-view stereo. 
Our Unification can 
recover finer 3D scene benefits from the direct learning cost volume,
and UFL is able to capture more fine-grained indicators for rebalancing 
samples and deal with continuous labels more reasonably. What's more valuable is that 
these two modules 
don't impose any memory or 
computational costs. Each plug-and-play module can be easily integrated into the existing 
MVS framework and achieve significant performance improvements, and we have shown this 
through 
our UniMVSNet. In the future, we plan to explore the integration of our modules into 
stereo matching or monocular field and look for more concise loss functions.
\\[5pt]
{\bf Acknowledgements.} Thanks to the National Natural Science Foundation of China 
62072013 and U21B2012, Shenzhen Research Projects of JCYJ20180503182128089 and 
201806080921419290, Shenzhen Cultivation of Excellent Scientific and Technological 
Innovation Talents RCJC20200714114435057,Shenzhen Fundamental Research Program 
(GXWD20201231165807007-20200806163656003). In addition, we thank the
anonymous reviewers for their valuable comments.

{\small
\bibliographystyle{ieee_fullname}
\bibliography{egbib}
}

\clearpage

\renewcommand{\thesection}{\Alph{section}}
\renewcommand{\thetable}{\Alph{table}}
\renewcommand{\thefigure}{\Alph{figure}}
\renewcommand{\theequation}{\alph{equation}}

{\centering \section*{Supplementary Material}}

\setcounter{section}{0}
\setcounter{figure}{0}
\setcounter{table}{0}
\setcounter{equation}{0}

\section{More Explanation of Unified Focal Loss}
\label{sec:meufl}

As shown in Equation (9), the dedicated function to control the range of scaling factor is 
designed as the sigmoid-like function as:
\begin{equation}
  S_b(x) = \frac{1}{(1 + b^{-x})}
\label{eq:df}
\end{equation}
where $x=\frac{|q-u|}{q^+}$ in this paper and its range is $[0, +\infty)$, therefore, 
the range of $S_b(x)$ is $[0.5, 1)$. As aforementioned, we adopt an asymmetrical scaling 
strategy to protect the precious positive learning signals and scale the range of 
$S_5^+$ to $[1,3)$ and $S_5^-$ to $[0,1)$. The detailed implementation of $S_5^+$ is:
\begin{equation}
  S_5^+(x) = 4 \times (\frac{1}{1 + 5^{-x}} - 0.5) + 1
\label{eq:df_pos}
\end{equation}
and the detailed implementation of $S_5^-$ is:
\begin{equation}
  S_5^-(x) = 2 \times (\frac{1}{1 + 5^{-x}} - 0.5)
\label{eq:df_neg}
\end{equation}

\section{Finer DTU Ground-truth}
\label{sec:fgt}

As mentioned in our main paper, we adopt the finer ground-truth to train our model 
additionally for a fair comparison with the start-of-the-art methods \cite{wei2021aa}. 
The refinement of each DTU ground-truth is achieved by cross-filtering with its 
neighbor viewpoints. For convenience, we directly adopt the processed results 
provided in \cite{wei2021aa}, and we only adopt the mask which indicates the 
validity of each point. Concretely, we adopt the union of the mask provided in 
\cite{gu2020cascade,yao2018mvsnet} and the up-sampled mask provided in \cite{wei2021aa} 
as the final mask.

\section{More Ablation Studies on DTU Dataset}
\label{sec:masdd}

Here, we perform more ablation studies on DTU to show you more information 
about our implementation.

\begin{figure}[h]
  \begin{center}
     \includegraphics[trim={3.5cm 5.5cm 3.5cm 1.9cm},clip,width=1.0\linewidth]{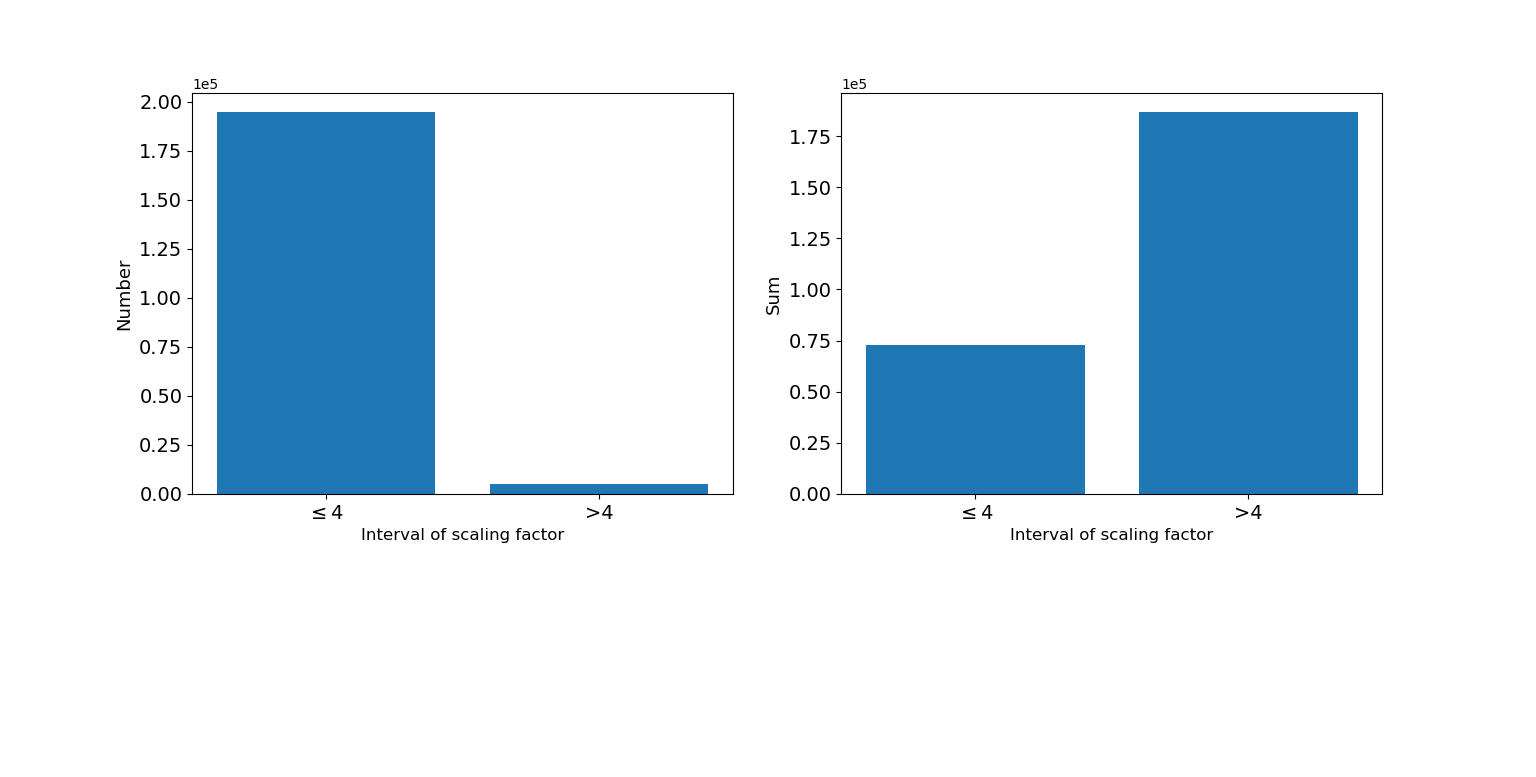}
  \end{center}
  \caption{{\bf The statistics of scaling factor $x=\frac{|q-u|}{q^+}$ in \cref{eq:ufl_naive}.}}
  \label{fig:ufl_stat}
\end{figure}

\noindent
{\bf The scaling factor in UFL.} The range of scaling factor in \cref{eq:ufl_naive} 
is $[0, +\infty)$. We count the average number and sum of scaling factors that fall in 
different intervals. As shown in \cref{fig:ufl_stat}, most of the scaling factors 
are less than 4 (Left figure). Even though, those 
small fractions of larger scaling factors take more weight (Right figure). This results 
in those abnormally large scaling factors occupying the model's training, and lead to 
difficulty in model convergence and extremely poor performance. Therefore, we introduce 
a dedicated function to control the scaling factors' range.

\begin{figure}[t]
  \begin{center}
     \includegraphics[trim={2.5cm 0.5cm 3cm 1.5cm},clip,width=1.0\linewidth]{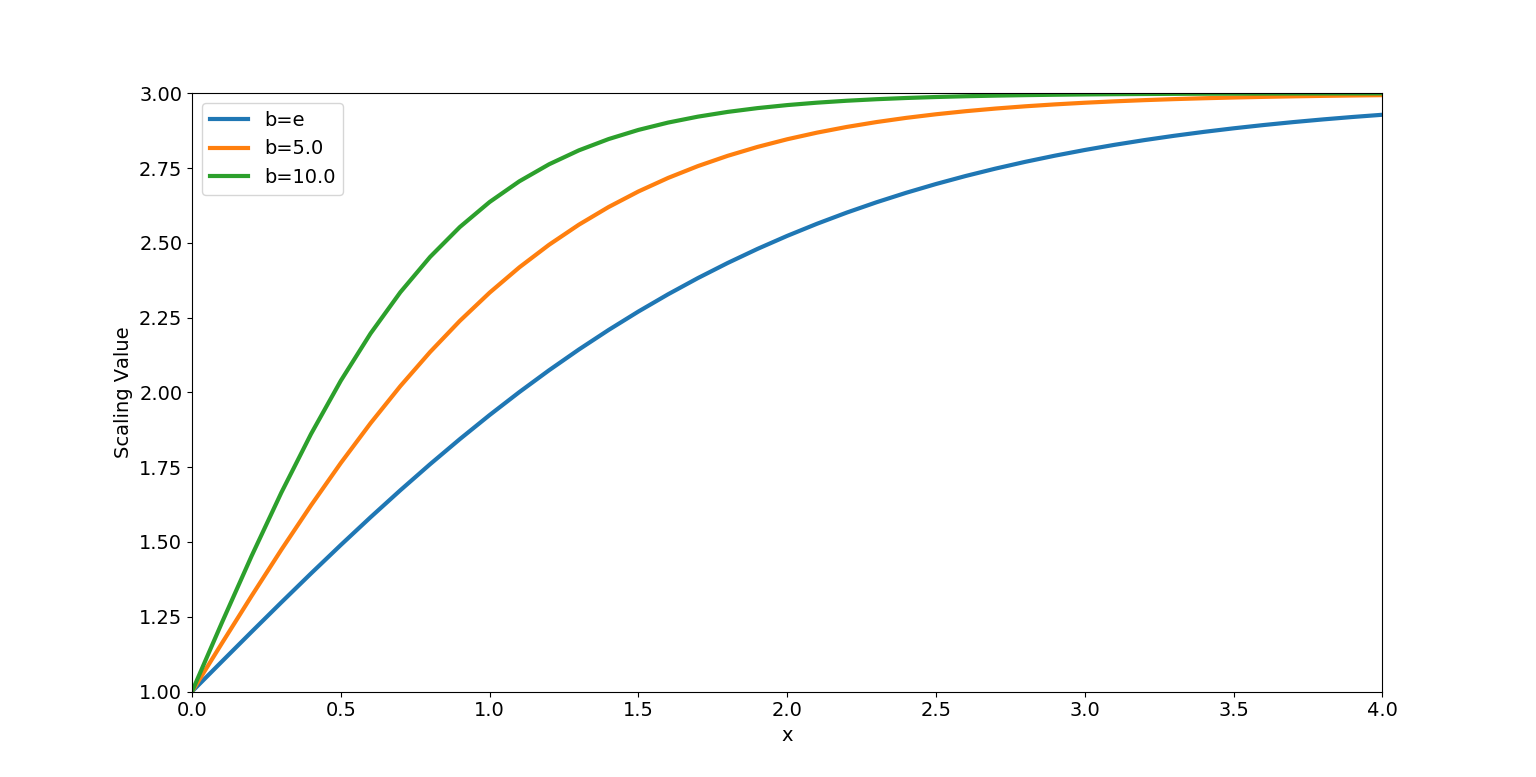}
  \end{center}
  \caption{{\bf The scaling value obtained through the positive dedicated function under 
  different base numbers.}}
  \label{fig:base_number}
\end{figure}

\noindent
{\bf The dedicated function in UFL.} As described in our main paper, we design the 
positive dedicated function as \cref{eq:df_pos} and negative dedicated function 
as \cref{eq:df_neg}. 
In fact, this final implementation is confirmed under our more experimental results. 
As shown in \cref{tb:df}, compared to adopting a common {\em sigmoid} function with 
a base $e$, it's better to use a dedicated function with a base 5. Meanwhile, scaling the 
range of dedicated function to $[1,3)$ is better than $[1,2)$. As shown in 
\cref{fig:base_number}, the base number controls the speed at which the function converges 
to the maximum value. The smaller the base number, the slower the convergence. In our 
experiments, we found that most of the points whose  $x=\frac{|q-u|}{q^+}$ is in the 
interval $[0, 4]$, so the scaling value calculated by the dedicated function needs to be 
distinguishable for the points in this interval, and we set $b=5$ in this paper. To 
be honest, we have only conducted a limited number of the base number the 
range of dedicated function as shown in \cref{tb:df} due to the time and resource 
considerations, and we believe there will be more powerful configurations.

\begin{table}[h]
  \begin{center}
  \resizebox{1.0\linewidth}{!}{
  \begin{tabular}{c|c|c|c|c}
  \hline
  Base Number & Range & ACC.(mm) & Comp.(mm) & Overall(mm) \\
  \hline
  $e$ & $[1, 2)$ & 0.354 & 0.282 & 0.318 \\
  \hline
  5 & $[1, 2)$ & 0.354 & 0.280 & 0.317 \\
  \hline
  5 & $[1, 3)$ & 0.352 & 0.278 & 0.315 \\
  \hline
  \end{tabular}
  }
  \end{center}
  \caption{{\bf Ablation results of dedicated function.} While the base number is ablated 
  for both the positive and negative dedicated function, we only ablate the range of 
  positive dedicated function.}
  \label{tb:df}
\end{table}

\noindent
{\bf The tunable parameter in UFL.} Tunable parameters in UFL like $\alpha$ and $\lambda$ 
are also important for rebalancing samples. As mentioned in our main paper, we always set 
$\alpha^+=1$ to protect the positive learning signals and configure 
other tunable parameters stage by stage due to the different number of depth hypotheses. 
In our implementation, we set the number of depth hypotheses to 48, 32, and 8 from stage1 to 
stage3. Apparently, the sample imbalance in stage1 is the most challenging, while it's 
the most relaxing or even negligible in stage3. As shown in \cref{tb:tp}, applying the same 
configuration across all stages performs the worst which indicates that the 
imbalances faced by different stages are different.

\begin{figure}[t]
  \begin{center}
     \includegraphics[trim={10cm 6.9cm 12cm 9cm},clip,width=1.0\linewidth]{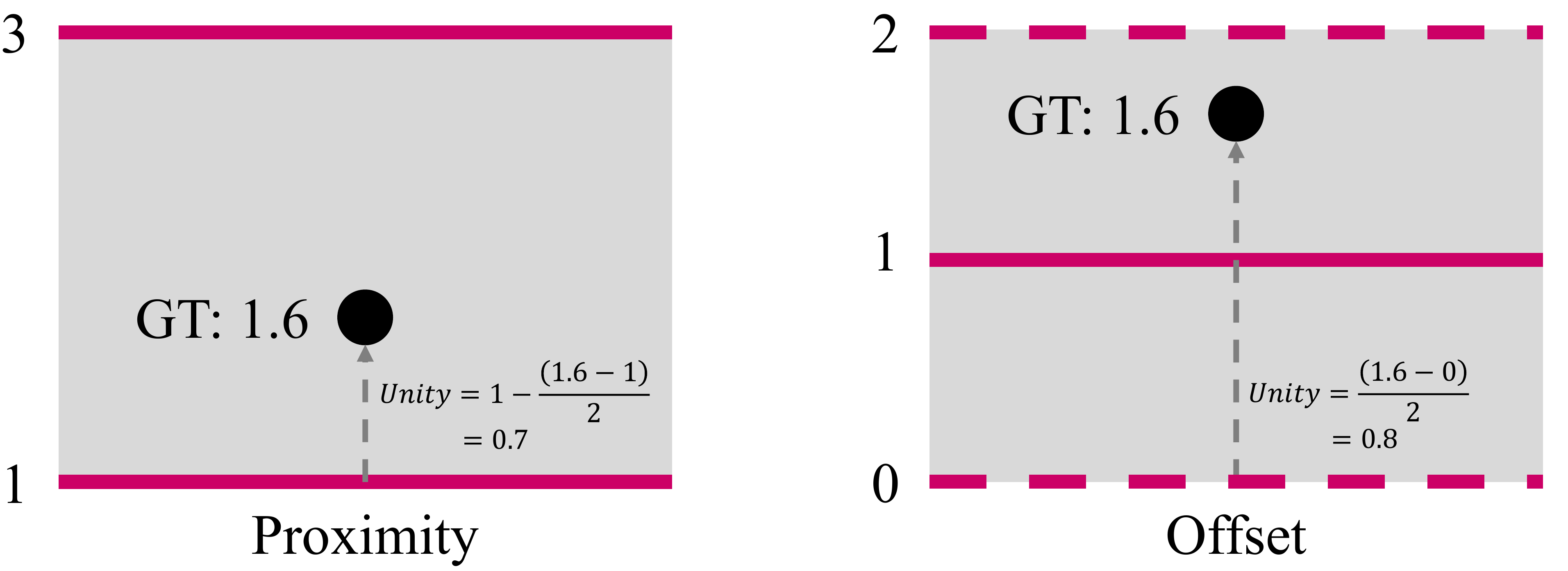}
  \end{center}
  \caption{{\bf Two available solutions to generate {\em Unity} (supervised signal) from the 
  ground-truth depth.} For proximity, we use the interval above the depth hypothesis 
  as its regression interval, and for offset, use the area where the depth hypothesis 
  is the median value as the regression interval.}
  \label{fig:proximity_offset}
\end{figure}

\noindent
{\bf Proximity $\mathcal{VS}$. Offset.} Different from the {\em Regression} and 
{\em Classification}, we propose {\em Unification} to classify the optimal depth hypothesis 
and regress its offset to ground-truth depth simultaneously. As shown in 
\cref{fig:proximity_offset}, there are two ways to regress the offset. The first is to 
predict proximity which is the complement of the offset and is also the method we adopt in 
this paper. The second is to directly estimate the offset. The comparison results 
between them are shown in \cref{tb:proximity_offset}. We can see that adopting proximity 
to regress the offset indirectly is much more powerful than purely using offset. As mentioned 
in our main paper, our {\em Unification} can be decomposed into two parts: classify the 
optimal depth hypothesis first and then regress the proximity for it. We infer that the 
reason why proximity is better than offset is the positive relationship between the 
magnitude of proximity and the quality of the classified optimal depth hypothesis in 
the first step. Meanwhile, 
we think that there should be better settings to improve the performance of using offset, 
but we have not made more attempts here.

\begin{table}[t]
  \begin{center}
  \resizebox{1.0\linewidth}{!}{
  \begin{tabular}{ccc|ccc|c|c|c}
  \hline
  \multicolumn{3}{c|}{$\alpha^-$} & \multicolumn{3}{c|}{$\lambda$} & \multirow{2}*{ACC.(mm)} & \multirow{2}*{Comp.(mm)} & \multirow{2}*{Overall(mm)} \\
  \cline{1-6}
  stage1 & stage2 & stage3 & stage1 & stage2& stage3 & & &\\
  \hline
  0.75 & 0.75 & 0.75 & 2 & 2 & 2 & 0.347 & 0.325 & 0.336 \\
  \hline
  1 & 1 & 1 & 2 & 1 & 0 & 0.366 & 0.282 & 0.324 \\
  \hline
  0.75 & 0.50 & 0.25 & 2 & 1 & 0 & 0.353 & 0.287 & 0.320 \\
  \hline
  \end{tabular}
  }
  \end{center}
  \caption{{\bf Ablation results of tunable parameters.} These experiments are conducted on 
  the model with only our {\em Unification} and UFL, and don't adopt the finer DTU 
  ground-truth or adaptive aggregation.}
  \label{tb:tp}
\end{table}

\begin{table}[t]
  \begin{center}
  \resizebox{1.0\linewidth}{!}{
  \begin{tabular}{c|c|c|c}
  \hline
  Method & ACC.(mm) & Comp.(mm) & Overall(mm) \\
  \hline
  Offset & 0.429 & 0.336 & 0.383 \\
  \hline
  Proximity & 0.372 & 0.282 & 0.327 \\
  \hline
  \end{tabular}
  }
  \end{center}
  \caption{{\bf Comparison of proximity and offset.} These experiments are conducted on 
  the model only using our {\em Unification}.}
  \label{tb:proximity_offset}
\end{table}

\section{More Comparisons between Unification, Classification and Regression}
\label{sec:meu}
{\bf (1)} Regression methods are harder to converge and have a greater risk of overfitting 
due to its indirect learning strategy, which has been studied in \cite{zhang2020adaptive}.
Meanwhile, they tend to generate smooth depth in object 
boundaries, because they treat the depth as the expectation of the depth hypotheses. However, 
they can achieve sub-pixel depth estimation, therefore, they have better accuracy. 
{\bf (2)} Classification methods cannot generate accurate depth 
due to their discrete prediction, but they constrain the cost volume directly and achieve 
better completeness. 
{\bf (3)} Our unification is exactly the complement of these two approaches. 
{\em Take the essence, get rid of the dross}. On the one 
hand, We directly constrain the cost volume to keep the model robust 
and pick the regression interval with the maximum unity to maintain the sharpness of the 
object boundary. On the other hand, we regress the proximity in the picked regression 
interval to generate accurate depth. Therefore, our unification is hoped to achieve regression's 
accuracy and classification's completeness. The results shown in \cref{tb:ab} just 
prove this.

\section{More Results on DTU Dataset}
\label{sec:mrdd}

\begin{figure*}
  \begin{center}
     \includegraphics[trim={18.8cm 0.6cm 17.8cm 0.6cm},clip,width=1.0\linewidth]{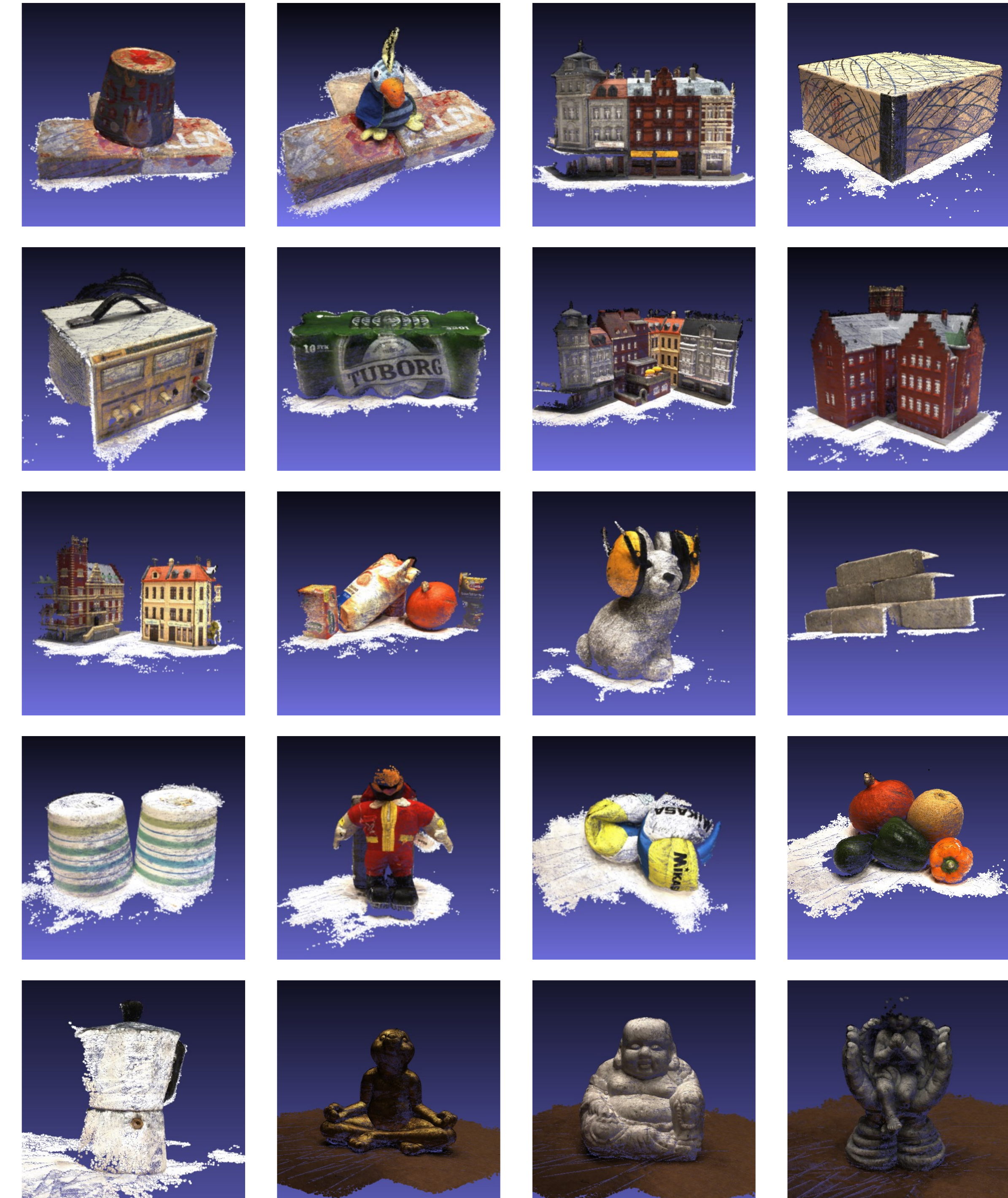}
  \end{center}
  \caption{{\bf More qualitative results on DTU dataset.}}
  \label{fig:sub_dtu_point}
\end{figure*}

\Cref{fig:sub_dtu_point} shows our additional point clouds reconstruction results. 
It can be seen that the point cloud reconstructed 
by our method has excellent accuracy and completeness.

\end{document}